\documentclass[sigconf, screen, nonacm]{acmart}

\makeatletter
\def\@affiliationfont{\@authorfont}
\def\@mkabstract{\bgroup
  \ifx\@abstract\@lempty\else
  {\phantomsection\addcontentsline{toc}{section}{\abstractname}
    \if@ACM@journal
       \everypar{\setbox\z@\lastbox\everypar{}}\small
    \else
      \section*{\abstractname}
    \fi
   \normalsize
   \ignorespaces\@abstract\par}
  \fi\egroup}
\makeatother

\usepackage{microtype}
\usepackage{graphicx}
\usepackage{subcaption}
\usepackage{booktabs}

\usepackage{hyperref}
\usepackage{pifont}
\usepackage{subcaption}
\usepackage{booktabs}

\usepackage[inline]{enumitem}

\usepackage{arydshln}
\usepackage[most]{tcolorbox}
\usepackage{multirow}

\usepackage[most]{tcolorbox}
\definecolor{darkolivegreen}{rgb}{0.33, 0.42, 0.18}
\definecolor{my_green}{RGB}{40,154,121}
\definecolor{my_yellow}{RGB}{255,165,0}
\definecolor{my_red}{RGB}{176,46,46}

\definecolor{red}{RGB}{184,49,55}

\definecolor{blue}{RGB}{55,83,156}

\definecolor{green}{RGB}{100,141,63}

\definecolor{titlered}{RGB}{184,49,55}
\newcommand{\titlered}[1]{\textcolor{titlered}{#1}}

\definecolor{vanblue}{RGB}{247,249,255}
\definecolor{vitgreen}{RGB}{245,251,247}

\definecolor{myblue}{RGB}{215,226,240}
\definecolor{mygreen}{RGB}{229,238,226}
\definecolor{mygrey}{RGB}{230, 230, 230}

\usepackage{algorithmic}
\usepackage{mathrsfs}
\usepackage{setspace}

\usepackage{mathtools}
\usepackage{amsthm}
\usepackage{adjustbox}

\usepackage[capitalize,noabbrev]{cleveref}

\theoremstyle{plain}

\theoremstyle{definition}

\theoremstyle{remark}

\usepackage[table]{xcolor}

\usepackage{mathtools}
\usepackage{booktabs}
\usepackage[inkscapelatex=false]{svg}
\usepackage{thmtools}
\usepackage{caption}
\usepackage{multirow}
\usepackage{float}

\AtBeginDocument{
  }

\begin{document}

\title[Thinking in Video: Can Video Generators Really Reason About the Real World?] {Thinking in Video: \\ Can Video Generators Really Reason About the Real World?}

\author{\vspace{1em}Yongheng Zhang$^{1,2*}$ \quad Guang Yang$^{1*}$ \quad Ruihan Hou$^{1*}$ \quad Qiguang Chen$^{1*}$ \quad Ziang Liu$^1$ \quad \\ Xiaolong Liu$^2$ \quad Manman Zhang$^2$ \quad Yanchao Hao$^2$ \quad Zheng Wei$^2$ \quad Hao Wu$^3$ \\ \quad Libo Qin$^1$ \quad Peishan Dai$^{1\dagger}$ \quad Yinghui Li$^3$ \quad Di Yin$^2$ \quad Xing Sun$^2$}
\affiliation{\vspace{1em}
  \institution{$^1$Central South University \quad
  $^2$Tencent \quad
  $^3$Tsinghua University}
  \country{}
}

\renewcommand{\shortauthors}{Zhang et al.}

\begin{abstract}
\begingroup
\renewcommand{\thefootnote}{\fnsymbol{footnote}}
\footnotetext[1]{Equal Contribution.}
\footnotetext[2]{Corresponding Author.}
\endgroup
Recent advances in world models and video generation have given rise to an emerging reasoning paradigm that leverages video generative models to simulate, predict, and reason about real-world dynamics. We redefine this paradigm as \textbf{Thinking in Video}, where video is not merely an output artifact but a medium for constructing, extending, and verifying causal thought. However, this promise remains unverified: convincing rollouts may reflect memorized appearances rather than causal understanding, while existing metrics separate perceptual fidelity from semantic logic. To evaluate whether video generators support such reasoning, we introduce the \textbf{Causal-Generative Dual-Judge} (\texttt{CGDJ}), auditing \textit{World Model Consistency} from two perspectives. \textbf{Explicit Causal Perception} tests whether a generator reads a video scenario as a reasoning problem through spatio-temporal flattened visual question answering, while \textbf{Implicit Generative Perception-Prediction Gap} evaluates whether it renders the causal consequence as a consistent future video. Applying \texttt{CGDJ} to representative open- and closed-source generators reveals a clear \textbf{Perception-Prediction Gap}: open-source models produce plausible dynamics despite near-zero explicit causal perception, whereas advanced closed-source systems show stronger but still limited alignment between reasoning and generation. Further analysis exposes audio-visual misalignment, where models verbalize correct causal logic more reliably than they render it, challenging the ``world simulator'' narrative.
\end{abstract}



\begin{teaserfigure}
	\centering
  \includegraphics[width=0.98\textwidth]{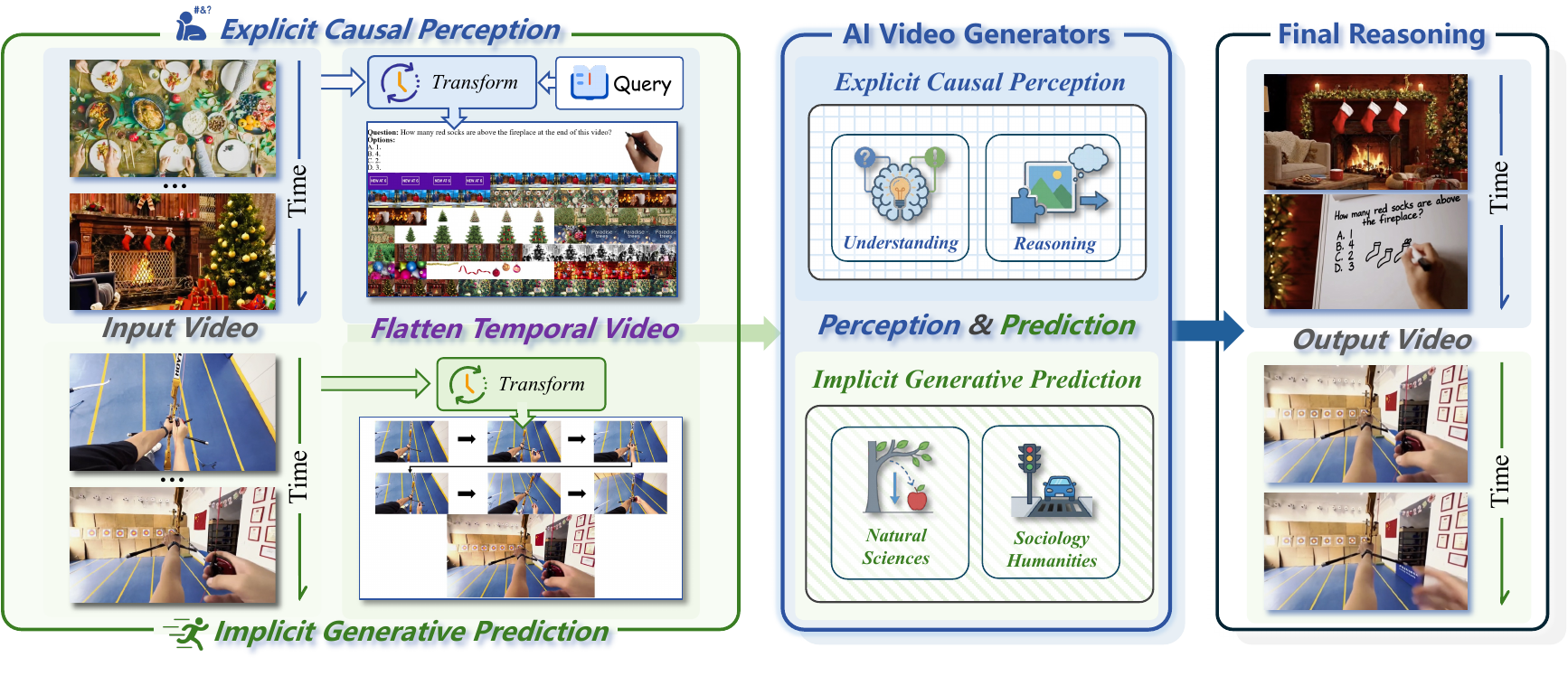}
  \caption{The overall workflow of the Causal Generative Dual Judge (\texttt{CGDJ}) consists of two components, \textbf{Explicit Causal Perception} (top) and \textbf{Implicit Generative Prediction} (bottom). Specifically, the pipeline transforms raw video sequences into model-compatible image inputs via the \textbf{Flatten Temporal Video} strategy. The video generator then processes this context to execute the assigned task and synthesize the final reasoning video. \vspace{1em}}
  \label{fig_intro}
\end{teaserfigure}

\maketitle

\section{Introduction}

Recent progress in video generation has changed what video models appear able to do. Autoregressive and diffusion-based systems now synthesize high-definition, minute-long clips with coherent objects, motions, and interactions~\citep{lu2026youtullmunlockingnativeagentic,blattmann2023stable,sora2,wu2025hunyuanvideo,veo3,wan2025wan,zhang2026chatbot}. This leap motivates a claim beyond visual synthesis: if a model can roll out plausible futures from visual conditions, it may use video as a medium for reasoning. We call this direction \textbf{Thinking in Video}: instead of merely describing answers in language, the model constructs, extends, and audits a visual world over time. Video generators thus become candidate \textbf{General World Simulators}, internalizing the physical laws and social dynamics that govern reality across diverse scenarios~\citep{ali2025world,ding2025understanding,bansal2025videophy}. Recent attempts at ``thinking with video'' explore this possibility for mathematics, navigation, and embodied problem solving~\citep{li2022past,tong2025thinking,yang2025reasoning,chow2025weave,chen2025tivibench,team2025evaluating,wei2025ggbench,liu2026let}.

\vspace{1mm}

However, the central promise of \textbf{Thinking in Video} remains unverified. A convincing rollout is not necessarily a correct reasoning process. A model may generate shattering glass, a falling object, or a human response because it understands causal conditions or memorized high-frequency visual associations~\citep{liu2024physgen,lin2025exploring}. The distinction matters: reasoning requires counterfactual and causal validity, not photorealistic continuation alone. Existing evaluation often rewards perceptual quality through Fréchet Video Distance (FVD) or Inception Score, while ignoring whether the generated future obeys scene mechanics~\citep{salimans2016improved,unterthiner2019fvd,ge2024content}. Without separating visual plausibility from causal correctness, treating video generators as reasoning engines for open-world decision making is premature~\citep{gupta2024essential,somepalli2023diffusion,valmeekam2023planbench,motamed2025generative,wu2026roboalignr1distilledmultimodalreward}.

\vspace{1mm}

Existing benchmarks rarely test whether a model can \textit{understand} a causal situation and \textit{act} through generation. This reflects a structural deficiency: paradigms bifurcate visual fidelity from semantic logic. Distributional metrics measure smoothness, realism, or text alignment, yet hallucinations may score well~\citep{huang2024vbench,liu2024evalcrafter,liu2024physgen,brooks2024video,le2025gravity}. Static multimodal benchmarks probe symbolic or visual question-answering, but not evolving video answers~\citep{sun2025t2v,rawte2025vibe}. Thus, when a generator seems to reason, we cannot tell whether it simulates causality or imitates appearances. This gap hides \textbf{Perception-Prediction Dissonance}: a model may know the outcome yet fail to render consistent pixels. Evaluating \textbf{Thinking in Video} requires auditing perception and prediction across physical and social dynamics~\citep{ha2018world,lecun2022path,li2025have}.

\vspace{1mm}

To close this gap, as shown in Figure~\ref{fig_intro}, we introduce the \textbf{Causal-Generative Dual-Judge} (\texttt{CGDJ}), a benchmark and evaluation framework for assessing whether video generative models can truly \textbf{Think in Video}. Rather than asking only whether the output looks good, \texttt{CGDJ} asks whether the model can use video as a reasoning medium: it must recognize causal structure from visual evidence and then ground that structure by generating a valid temporal consequence. The paradigm contains two complementary modules:


\begin{itemize}
\item [\ding{182}] \textbf{Explicit Causal Perception}: We test whether a model can identify and articulate a video's causal structure, rather than merely generate a plausible continuation. Since most video generators cannot process interleaved video-text prompts, our \textbf{\textit{Flatten Temporal Video}} strategy converts video frames and rasterized queries into a unified image grid. This creates an explicit, directly judgeable visual QA task, forcing the model to reveal its recognized causal progression instead of relying on statistical continuation.
    \item [\ding{183}] \textbf{Implicit Generative Prediction}: We test whether a model can enact its reasoning in video. Conditioned on a causal antecedent, the model must synthesize the consequent future sequence. This generative protocol forces the model to move beyond choosing or verbalizing an answer: it must physically render the outcome, enabling direct inspection of whether its visual rollout is causally valid.
\end{itemize}

Experiments reveal that current video generators are only partially capable of \textbf{Thinking in Video}. On the explicit side, advanced closed-source generators such as \texttt{Sora-2} and \texttt{Veo-3.1} show measurable causal perception, whereas open-source generators nearly collapse under the same flattened reasoning protocol. On the generative side, however, open-source models can still produce moderately plausible causal continuations, exposing a sharp \textbf{Perception-Prediction Gap}: visual action can surpass explicit causal understanding. Further analysis shows an audio-visual misalignment where models often articulate the correct causal logic more reliably than they render it. These results challenge a simple ``world simulator'' narrative. Current systems can imitate parts of world dynamics, and the strongest models begin to align perception with prediction, but robust video-based reasoning still requires stronger grounding between causal knowledge and visual computation.

\vspace{1mm}

Our key contributions are summarized as follows:
\begin{itemize}
    \item [(1)] We redefine reasoning with video generative models as the \textbf{Thinking in Video} paradigm, where video becomes a medium for constructing and verifying causal thought, and highlight the need to properly evaluate whether current generators can support such reasoning.
    \vspace{2mm}
    \item [(2)] We propose the Causal-Generative Dual-Judge (\texttt{CGDJ}) framework, with the Flatten Temporal Video strategy, to jointly evaluate \textit{Explicit Causal Perception} and \textit{Implicit Generative Prediction} in current video generative models.
    \vspace{2mm}
    \item [(3)] Through experiments, we reveal a fundamental Perception-Prediction Gap: current models can exhibit plausible visual dynamics without reliable causal perception, while advanced systems show early but still limited alignment between explicit reasoning and generative simulation.
\end{itemize}

To facilitate further research, all code and data will be publicly available: \url{https://github.com/BRZ911/Thinking-in-Video}.

\begin{figure*}[t]
	\centering
  \includegraphics[width=0.98\textwidth]{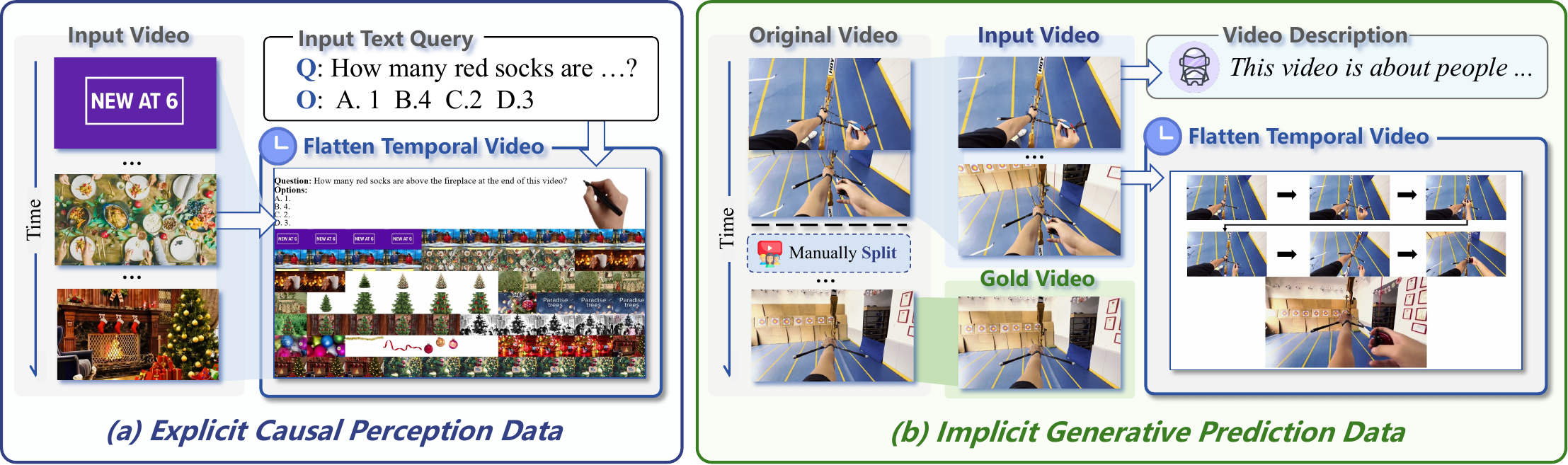}
  \caption{The construction process of the \texttt{CGDJ} benchmark. (a) \textbf{\textit{Explicit Causal Perception Data}} utilizes a \textit{Flatten Temporal Video} strategy to map temporal dynamics into a spatial grid, combining visual history with rasterized text queries for direct reasoning. (b) \textbf{\textit{Implicit Generative Prediction Data}} evaluates simulation by splitting videos at a critical inflection point; The pre-event frames serve as a flattened visual condition, prompting the model to generate the physically consistent post-event future.}
  \label{fig_data}
\end{figure*}

\section{Causal–Generative Dual-Judge}
\label{sec:cgdj}

To evaluate whether video generative models are simply mimicking pixel statistics or truly simulating world dynamics, we propose the \textbf{Causal–Generative Dual-Judge (\texttt{CGDJ})}. As shown in Figure~\ref{fig_data}, this paradigm decouples the evaluation of semantic comprehension (\textit{\textbf{Perception}}) from physical rendering (\textit{\textbf{Prediction}}). 
 \texttt{CGDJ} consists of: (i) \textit{\textbf{Causal–Generative Dual-Judge Benchmark}}~($\S \ref{sec:benchmark}$), and (ii) \textit{\textbf{Causal–Generative Dual-Judge Evaluation}}.

\subsection{Causal–Generative Dual-Judge Benchmark}
\label{sec:benchmark}
\subsubsection{Explicit Causal Understanding Data}

As shown in Figure~\ref{fig_data}~(a), the comprehensive benchmark construction process consists of two main phases: \textit{Data Collection} and \textit{Flatten Temporal Video Transition}:

\textbf{Data Collection.}
To evaluate the causal reasoning of video generative models, we align our benchmark with video understanding standards by adopting all 900 videos from Video-MME~\cite{fu2025video}. To ensure a comprehensive evaluation across different temporal scales, the collection spans short, medium, and long durations. Furthermore, the dataset covers a diverse range of real-world dynamics, categorized into six domains: (i)~\textit{Knowledge}, (ii)~\textit{Life Record}, (iii)~\textit{Sports Competition}, (iv)~\textit{Artistic Performance}, (v)~\textit{Film \& Television}, and (vi)~\textit{Multilingual}. The statistical distribution across these individual domains is presented in Figure~\ref{fig_data_num}~(a). This enables evaluating causal internalization across diverse physical and social dynamics.

\textbf{Flatten Temporal Video Transition.}
Since current video generative models are strictly restricted to processing only static image inputs, to overcome this fundamental limitation, we propose the \textbf{Flatten Temporal Video} strategy that maps a video-question pair onto a single static image $I_{\text{composite}} \in \mathbb{R}^{720 \times 1280 \times 3}$.

Formally, given a video $\mathcal{V}$ and a question-option pair $\mathcal{Q}$, the construction process consists of three steps:

\noindent \textbf{1. Temporal Grid Construction:} We extract a sequence of $N=70$ frames through uniform temporal sampling. These frames are concatenated into a spatial matrix $I_{\text{grid}}$ with dimensions $7 \times 10$:
\begin{equation}
    I_{\text{grid}} = \underset{r=1}{\overset{7}{\oplus}} \left( \underset{c=1}{\overset{10}{\oplus}} x_{t_{(r-1)10 + c}} \right)
\end{equation}
where $\oplus$ denotes spatial concatenation. This grid effectively transforms the temporal sequence into a spatial texture accessible to the visual encoder.

\noindent \textbf{2. Semantic Rendering:} Instead of relying on conventional pre-trained external text encoders, we directly transform the specific textual query $\mathcal{Q}$ into a dense pixel-space representation $I_{\text{text}}$ via a dedicated high-fidelity rasterization function. This newly rendered text block is strategically positioned to occupy the entire upper spatial visual field of the final image grid.

\noindent \textbf{3. Vertical Integration:} The final composite input for the model is meticulously synthesized by vertically concatenating the specific semantic instruction tensor and the flattened temporal grid tensor along the primary height dimension. In order to ensure full technical compatibility with existing standard video generation architectures, this newly fused tensor subsequently undergoes an adaptive resizing operation to match the required target resolution:
\begin{equation}
    I_{\text{composite}} = \text{Resize}_{1280 \times 720}([I_{\text{text}}; I_{\text{grid}}])
\end{equation}
This layout creates a ``visual representation'' by stacking text above the grid, enabling causal reasoning through joint attention to instructions and history.

\subsubsection{Implicit Generative Action Data}
As shown in Figure~\ref{fig_data} (b), the construction process consists of three main steps: \textit{Data Collection}, \textit{Video Splitting}, and \textit{Flattening of Temporal Video Transitions}:

\paragraph{Data Collection.}
We systematically classify the broad spectrum of World Dynamics into \textbf{Natural Sciences} and \textbf{Sociology \& Humanities}, balancing the evaluation between rigid physical mechanics and complex human behaviors.

\begin{itemize}
\vspace{2mm}
    \item \textbf{Natural Sciences:} This category evaluates the model's fundamental understanding of physical laws. We aggregate data from various authoritative benchmarks including \textit{The Sound of Water}~\citep{bagad2025sound}, \textit{Physion}~\citep{bear2021physion}, and \textit{Physics IQ Benchmark}~\citep{motamed2025generative}. We classify these into four fine-grained sub-domains:
(i)~\textit{Physical Mechanics}, (ii)~\textit{Natural \& Life Sciences}, (iii)~\textit{Materials \& Structures}, and (iv)~\textit{Fields \& Energy}.
\vspace{2mm}
    \item \textbf{Sociology \& Humanities:} This category evaluates the simulation of human behaviors and social logic. Data sources include the \textit{TLD Benchmark} (Vehicle Signals)~\citep{chai2024tld}, \textit{Kinetics Human Action}~\citep{kay2017kinetics}, \textit{Ego4D}~\citep{grauman2022ego4d}, and selected web videos. These are categorized into sub-domains: (i)~\textit{Transportation Behavior}, (ii)~\textit{Life Behavior}, (iii)~\textit{Occupational Role Behavior}, (iv)~\textit{Public Social Behavior}, and (v)~\textit{Sports Activity Behavior}.
\vspace{2mm}
\end{itemize}

\begin{figure}[t]
	\centering
  \includegraphics[width=0.95\linewidth]{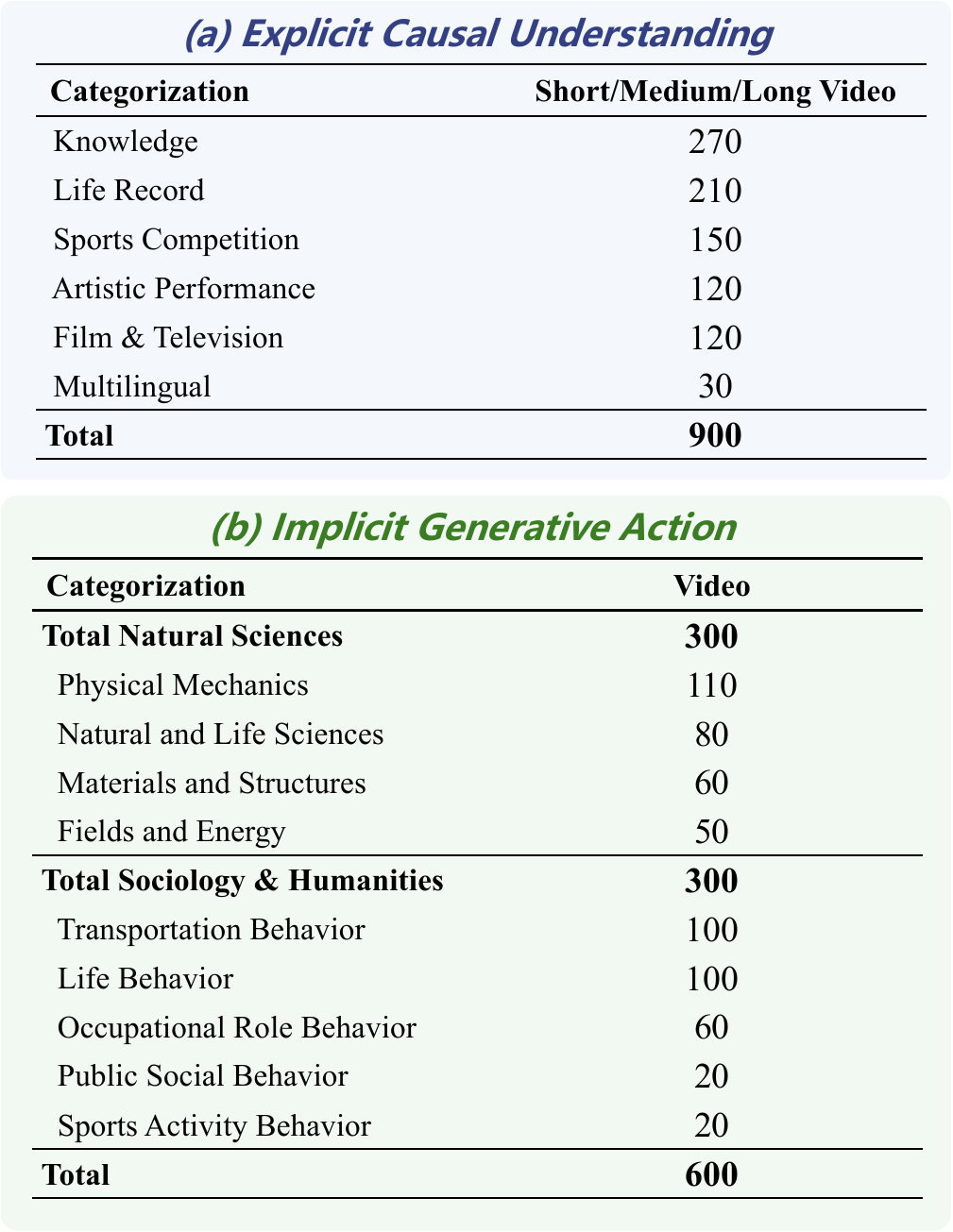}
  \caption{Statistical distribution of the Causal-Generative Dual-Judge Benchmark. (a) The distribution of 900 videos for Explicit Causal Understanding across six domains. (b) The classification of 600 videos for Implicit Generative Action.}
  \label{fig_data_num}
\end{figure}

As shown in Figure~\ref{fig_data_num} (b), the final curated benchmark comprises a total of 600 high-quality videos. To ensure an unbiased evaluation between physical rigidity and social complexity, we maintain a strictly balanced distribution across the two domains: \textit{300 samples for Natural Sciences} and \textit{300 samples for Sociology \& Humanities}.

\paragraph{Video Splitting.}
To rigorously reformulate the standard generative task as a structured causal prediction problem, we implement a precise temporal segmentation strategy. For every video in the benchmark, expert annotators manually identify and pinpoint the \textbf{Causal Inflection Point} (or critical node). This specific timestamp represents the definitive boundary where the accumulation of causal conditions triggers an irreversible event or state transition. 

\begin{itemize}
\vspace{2mm}
\item [\ding{182}] \textbf{Input Video:} The pre-event sequence serves as the visual condition, providing the necessary causal context for the subsequent generation. \textit{Furthermore, we use \texttt{Gemini-3-Pro} \citep{team2025gemma} to produce a detailed semantic video description, which is co-injected with the video to facilitate the generation.} 
\vspace{2mm}
\item [\ding{183}] \textbf{Gold Video:} The subsequent temporal segment functions as the immutable ground truth reference, capturing the realized outcome of the initial conditions. This empirically recorded footage serves as a rigorous baseline to audit the validity of the model's extrapolation.
\vspace{2mm}
\end{itemize}

\paragraph{Flattening of Temporal Video Transitions.}
Similar to the previously mentioned strategy in the Explicit Understanding module, we employ a flattening technique to input the video context into the generative model. Specifically, we extract $N=7$ keyframes from the \textit{Input Video} at uniform discrete temporal steps to represent the motion trajectory. Formally, these frames are spatially concatenated to form a continuous visual condition $I_{\text{motion}}$:
\begin{equation}
    I_{\text{motion}} = \underset{i=1}{\overset{N}{\oplus}} x_{t_i}
\end{equation}
where $x_{t_i}$ represents the frame at time step $t_i$ and $\oplus$ denotes spatial horizontal concatenation. This arrangement constructs a coherent ``visual arrow of time'', effectively encoding the directional vector of motion and dynamic temporal state changes.

\subsection{Causal–Generative Dual-Judge Evaluation}
\label{sec:evaluation}

\begin{figure*}[t]
	\centering
  \includegraphics[width=0.98\textwidth]{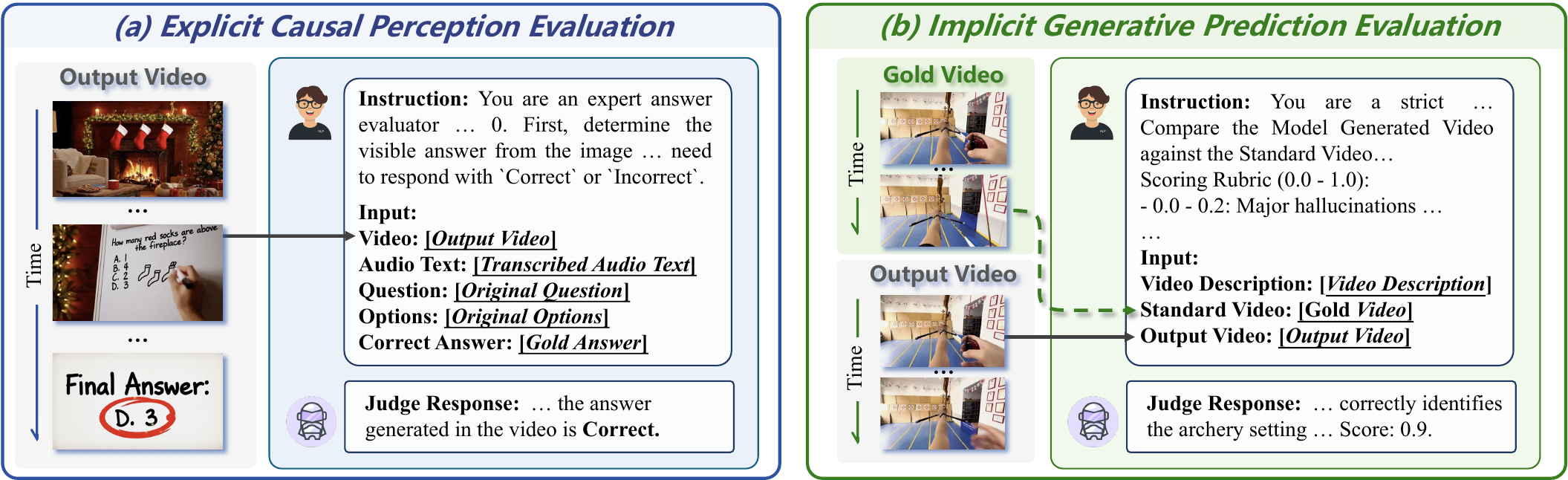}
  \caption{The overall evaluation pipeline of the Causal–Generative Dual-Judge consists of two distinct yet complementary protocols: \textit{(a) Explicit Causal Understanding Evaluation} and \textit{(b) Implicit Generative Action Evaluation}. }
  \label{fig_eval}
\end{figure*}

To quantify the capabilities of the model in simulating world dynamics, as shown in Figure~\ref{fig_eval}, we design an automated evaluation pipeline based on advanced multimodal models.

\subsubsection{Explicit Causal Evaluation}

To assess causal validity through reasoning, as shown in Figure~\ref{fig_eval} (a), we establish a robust, comprehensive, and fully automated two-stage framework: \textbf{Audio-Text Alignment} and \textbf{MLLM-based Causal Adjudication}.
\paragraph{Audio Context Extraction.}
The generated videos often contain the correct solution or reasoning steps encoded directly within the audio. To strictly evaluate this verbal information, we utilize the \texttt{whisper-large-v2}~\citep{radford2023robust} model to extract and transcribe the audio track. By converting the spoken problem-solving process into text, we ensure that the judge considers both the visual outcome and the model's narrated logic for a more holistic assessment.

\paragraph{MLLM-based Adjudication.}
We employ the \texttt{Gemini-3-Pro}~\citep{team2025gemma} as the core evaluator, owing to its superior video understanding and long-context reasoning capabilities critical for our analysis. 
Specifically, the evaluation process functions as follows:
\begin{enumerate}
\item [\ding{182}] \textbf{Input Construction:} We construct a composite prompt for Gemini, which includes the following key inputs: (1)~the \textit{Generated Video} produced by the model under test, (2)~the \textit{Transcribed Audio Text}, (3)~the \textit{Original Question and Option}, and (4)~the \textit{Ground Truth Answer}.
\vspace{1mm}
\item [\ding{183}] \textbf{Causal Judgment:} \texttt{Gemini-3-Pro} is tasked to act as a judge. It compares the visual events in the generated video against the logical requirements of the correct answer. The MLLM determines whether the generated video depicts the causal consequence implied by ground truth, classifying the generation as \textit{``Correct''} or \textit{``Incorrect''}. We validate this automated metric against human judgment in~Figure~\ref{fig_human_align}.
\end{enumerate}
This automated pipeline ensures that the evaluation moves beyond superficial pixel metrics and focuses on the semantic and causal correctness of the generated content.

\subsubsection{Implicit Generative Evaluation}

To evaluate the real-world simulation authenticity, as shown in Figure~\ref{fig_eval} (b), we employ a comprehensive, systematic, and rigorous \textit{Reference-Based Comparative Evaluation}. Using \texttt{Gemini-3-Pro} effectively serves as a dedicated ``Video Quality Auditor''. It systematically benchmarks the video generated by the model against the ground truth ``Gold Video'', while ensuring strict adherence to the governing textual description.

\paragraph{Scoring Mechanism.}
The automated evaluator assigns a score $S \in [0.0, 1.0]$ based on the following three dimensions: (i)~\textit{Semantic Alignment} (object/action correctness), (ii)~\textit{Reference Consistency} (logical flow vs. ground truth), and  (iii)~\textit{Physical Validity} (absence of flickering/physics violations). Scores range from $0.0$ (hallucinations) to $1.0$ (perfect match), effectively quantifying the model's innate ability to ground reasoning into physical action.

\section{Experiments}

In this section, we conduct a comprehensive empirical analysis to validate the newly proposed Causal–Generative Dual-Judge (\texttt{CGDJ}). Our systematic experiments are structured to probe the two most distinct facets of world simulation:
\begin{itemize}
\vspace{1mm}
    \item [(1)] \textbf{Explicit Causal Understanding:} We first assess whether various models possess the essential theoretical knowledge to identify correct causal dynamics governing complex real-world events through specific analytical tasks.
\vspace{1mm}
    \item [(2)] \textbf{Implicit Generative Action:} We then evaluate the intrinsic ability of the video generation models to physically ground this knowledge by generating valid future video sequences that closely align with real-world dynamics.
 \vspace{1mm}
\end{itemize}
By comparing these two complementary modalities, we aim to further quantify the ``Perception-Prediction Gap'' and diagnose the state of modern video generative models as world simulators.

\subsection{Experimental Setup for Explicit Causal Understanding}
\label{sec:exp_explicit_setup}

To evaluate the explicit reasoning capabilities of video generative models, we benchmark them against advanced MLLMs to quantitatively highlight the critical perception and reasoning performance gap. The experimental setup is designed as follows:

\paragraph{Multimodal Large Language Model Baselines.}
To establish a performance ceiling and validate the effectiveness of our visual representation, we evaluate a suite of advanced MLLMs. This includes \textit{Qwen3-VL-Instruct} (2B, 4B, and 8B)\citep{bai2025qwen3vltechnicalreport}, as well as proprietary frontier models including \textit{GPT-5}\citep{singh2025openai} and \textit{Gemini-3-Flash}~\citep{team2025gemma}.
For these MLLMs, we design two distinct input protocols to analyze the impact of information formatting: \textbf{(1)~Setting~A~(Native Video Input):} The models process the original video tokens directly, utilizing their native temporal encoders.
\textbf{(2)~Setting~B~(Flatten Temporal Video Input):} The models process the ``flattened'' spatial grid image (as described in $\S \ref{sec:benchmark}$), identical to the input format used for video generative models. This comparison verifies whether the flattening strategy preserves sufficient causal information for reasoning. 

\begin{table}[t]
\caption{Detailed configurations of the representative baseline video generative models evaluated in our experiments.}
\setlength{\tabcolsep}{3pt} 
\centering 
\begin{adjustbox}{width=\linewidth} 
\begin{tabular}{lccccc}
\toprule 
\textbf{Model} & \textbf{Resolution} & \textbf{Source Type} & \textbf{Duration} & \textbf{Price} \\
\midrule 
\rowcolor[rgb]{ .970, .978, .999 }
Wan-2.2-14B~\citep{wan2025wan} & $1280 \times 720$ & open & 5s & - \\
\rowcolor[rgb]{ .970, .978, .999 }
Hunyuan-1.5~\citep{wu2025hunyuanvideo} & $1280 \times 720$ & open & 5s & - \\
\rowcolor[rgb]{ .970, .978, .999 }
Veo-3.1~\citep{veo3} & $1280 \times 720$ & closed & 8s & \$0.4/s \\
\rowcolor[rgb]{ .970, .978, .999 }
Sora-2~\citep{sora2} & $1280 \times 720$ & closed & 12s & \$0.1/s \\
\bottomrule 
\end{tabular}
\end{adjustbox}
\label{tab:model_specs}
\end{table}

\paragraph{Video Generative Model Baselines and Inference.}
We benchmark a representative set of leading video generation models, categorizing them into open-source and closed-source groups. Since models cannot natively process interleaved video-text inputs for understanding tasks, all are evaluated using our proposed \textit{Flatten Temporal Video} strategy. The configurations are tailored to model accessibility:

\begin{itemize} 
    \item \textbf{Open-Source Models (Local Inference):} We select \textit{Wan-2.2-14B}~\citep{wan2025wan} and \textit{HunyuanVideo-1.5}~\citep{wu2025hunyuanvideo}. We utilized their official implementations for inference on a local server.
\vspace{1mm}
    \item \textbf{Closed-Source Models (API Access):} We assess both advanced \textit{Veo-3.1}~\citep{veo3} and \textit{Sora-2}~\citep{sora2} via their official APIs. These models were tested using their default stable duration settings (up to 12s). To ensure economic transparency per benchmarks, we recorded pricing during evaluation.
\end{itemize}
Comprehensive parameter settings, including resolution, max duration, and pricing, are detailed for comparative purposes in Table~\ref{tab:model_specs}.


\begin{table*}[t]
\caption{Accuracy (\%) on \textit{Explicit Causal Understanding} Tasks. \raisebox{0.7mm}{\colorbox{mygrey}{\textcolor{mygrey}{\rule{0.7mm}{0.7mm}}}} Area: Setting~A~(Native Video Input).  \raisebox{0.7mm}{\colorbox{myblue}{\textcolor{myblue}{\rule{0.7mm}{0.7mm}}}} Area: Setting~B~(Flatten Temporal Video Input). The \titlered{Red} represents better AVG performance.}
\setlength{\tabcolsep}{7pt} 
\centering 
\begin{adjustbox}{width=0.99\textwidth}
\begin{tabular}{lccc|ccccccc}
\toprule 
\multirow{2.5}{*}{\textbf{Model}} & 
\multicolumn{3}{c}{\textbf{Video Length}} & 
\multicolumn{6}{c}{\textbf{Video Category}} & 
\multirow{2}{*}{\underline{\textcolor{titlered}{\textbf{AVG}}}} \\
\cmidrule(lr){2-4} \cmidrule(lr){5-10}
 & \textbf{Short} & \textbf{Medium} & \textbf{Long} & \textbf{Know.} & \textbf{Film.} & \textbf{Sports.} & \textbf{Artistic.} & \textbf{Life.} & \textbf{Multi.} & \\
\midrule 
\rowcolor[rgb]{ .980, .980, .980}
\multicolumn{11}{c}{\rule{0pt}{2ex}\textit{\fontsize{11pt}{11pt} Native Video Input (Language Model)} } \\
\midrule
\rowcolor[rgb]{ .980, .980, .980}
\text{Qwen3-VL-2B~\citep{bai2025qwen3vltechnicalreport}} & 70.3 & 52.0 & 49.0 & 63.7 & 59.2 & 48.7 & 55.0 & 56.7 & 43.3 & 57.1 \\
\rowcolor[rgb]{ .980, .980, .980}
\text{Qwen3-VL-4B~\citep{bai2025qwen3vltechnicalreport}} & 76.3 & 59.3 & 49.0 & 66.3 & 62.5 & 59.3 & 61.7 & 58.1 & 50.0 & 61.6 \\
\rowcolor[rgb]{ .980, .980, .980}
\text{Qwen3-VL-8B~\citep{bai2025qwen3vltechnicalreport}} & 82.3 & 65.3 & 56.7 & 71.5 & 73.3 & 62.7 & 66.7 & 66.7 & 60.0 & 68.1 \\
\rowcolor[rgb]{ .980, .980, .980}
\text{GPT-5~\citep{singh2025openai}} & 88.0 & 79.3 & 71.0 & 82.2 & 82.5 & 75.3 & 81.7 & 78.1 & 63.3 & 79.4 \\
\rowcolor[rgb]{ .980, .980, .980}
\text{Gemini-3-Flash~\citep{team2025gemma}} & 91.7 & 82.3 & 75.0 & 85.2 & 86.7 & 82.7 & 79.2 & 83.3 & 63.3 & \underline{\textcolor{titlered}{\textbf{83.0}}} \\
\midrule
\rowcolor[rgb]{ .970, .978, .999 }
\multicolumn{11}{c}{\rule{0pt}{2ex}\textit{\fontsize{11pt}{11pt} Flatten Temporal Video Input (Language Model)} } \\
\midrule
\rowcolor[rgb]{ .970, .978, .999 }
\text{Qwen3-VL-2B~\citep{bai2025qwen3vltechnicalreport}} & 48.7 & 40.0 & 42.3 & 47.0 & 41.7 & 40.0 & 40.0 & 45.7 & 40.0 & 43.7 \\
\rowcolor[rgb]{ .970, .978, .999 }
\text{Qwen3-VL-4B~\citep{bai2025qwen3vltechnicalreport}} & 57.3 & 40.3 & 43.7 & 54.8 & 46.7 & 38.0 & 41.7 & 48.1 & 40.0 & 47.1 \\
\rowcolor[rgb]{ .970, .978, .999 }
\text{Qwen3-VL-8B~\citep{bai2025qwen3vltechnicalreport}} & 64.7 & 50.7 & 50.0 & 61.5 & 56.7 & 49.3 & 47.5 & 53.8 & 60.0 & 55.1 \\
\rowcolor[rgb]{ .970, .978, .999 }
\text{GPT-5~\citep{singh2025openai}} & 74.3 & 65.0 & 63.3 & 72.2 & 73.3 & 58.7 & 68.3 & 65.2 & 60.0 & 67.6 \\
\rowcolor[rgb]{ .970, .978, .999 }
\text{Gemini-3-Flash~\citep{team2025gemma}} & 79.7 & 72.7 & 70.0 & 78.5 & 77.5 & 64.0 & 74.2 & 74.8 & 66.7 & \underline{\textcolor{titlered}{\textbf{74.1}}} \\
\midrule
\rowcolor[rgb]{ .970, .978, .999 }
\multicolumn{11}{c}{\rule{0pt}{2ex}\textit{\fontsize{11pt}{11pt} Flatten Temporal Video Input (Video Generative Model)} } \\
\midrule
\rowcolor[rgb]{ .970, .978, .999 }
\text{Wan-2.2-14B~\citep{wan2025wan}} & 1.0 & 0.7 & 0.3 & 0.4 & 0.8 & 0.7 & 1.7 & 0.5 & 0.0 & 0.7 \\
\rowcolor[rgb]{ .970, .978, .999 }
\text{HunyuanVideo-1.5~\citep{wu2025hunyuanvideo}} & 1.3 & 0.3 & 1.8 & 1.5 & 1.7 & 2.0 & 0.0 & 1.5 & 3.3 & 1.5 \\
\rowcolor[rgb]{ .970, .978, .999 }
\text{Veo-3.1~\citep{veo3}} & 52.3 & 48.0 & 49.3 & 54.1 & 49.2 & 48.3 & 44.6 & 51.9 & 30.0 & 49.9 \\
\rowcolor[rgb]{ .970, .978, .999 }
\text{Sora-2~\citep{sora2}} & 52.5 & 49.5 & 51.5 & 53.9 & 52.1 & 45.3 & 51.7 & 51.2 & 50.0 & \underline{\textcolor{titlered}{\textbf{51.2}}} \\
\bottomrule 
\end{tabular}
\end{adjustbox}
\label{tab:explicit_results}
\end{table*}

\subsection{Main Results}
Table~\ref{tab:explicit_results} and Table~\ref{tab:implicit_results} summarize the quantitative results of our Causal–Generative Dual-Judge evaluation. From these experimental comparisons, we observe that:

\vspace{2mm}

\textbf{1. Spatio-Temporal Flattening serves as a valid medium for causal reasoning.} 
To validate whether our flattening strategy preserves sufficient information for reasoning, we compare MLLM performance under Native versus Flatten Temporal Video Input. The results show that while transforming temporal videos into spatial grids naturally causes some information loss, advanced MLLMs maintain robust performance. Notably, \textit{Gemini-3-Flash} achieves \textit{\textit{74.1\%}} under the flattened setting, retaining approximately~\textit{\textit{89\%}} of its native performance~(\textit{\textit{83.0\%}}). Even the smaller \textit{Qwen3-VL-8B} maintains a respectable~\textit{81\%} relative to its native performance. This evidence confirms that the ``visual medium'' constructed by our flattening technique contains sufficient logical cues.

\vspace{2mm}

\textbf{2. Emergent but Limited Reasoning in Generative Models.}
A distinct divide emerges. Closed-source models such as \texttt{Sora-2} (\textit{51.2\%}) and \texttt{Veo-3.1} (\textit{49.9\%}) exhibit measurable explicit reasoning capabilities, performing on par with an 8B-parameter MLLM. This suggests that scaling video generation, or internal prompt rewriting pipelines, fosters an emergent, albeit rudimentary, understanding of world dynamics. However, they significantly trail behind state-of-the-art reasoning engines, Gemini-3-Flash at 74.1\%, indicating that current ``World Simulators'' are not yet robust reasoners. Notably, open-weights models like \texttt{Wan-2.2} (\textit{0.7\%}) and \texttt{HunyuanVideo} (\textit{1.5\%}) suffer a complete collapse in this discriminative task, likely due to a lack of instruction-following alignment in their training.

\vspace{2mm}

\textbf{3. Closed-Source Models Dominate Physical Simulation.} 
In the generative domain, \textbf{Sora-2} leads with an average score of \textbf{61.6\%}, followed closely by \texttt{Veo-3.1} (\textit{58.9\%}), demonstrating superior capability in simulating complex dynamics. Open-source models lag behind at \textit{42.5\%} and \textit{41.5\%} under our automated MLLM evaluation. Interestingly, models consistently perform better in Sociology \& Humanities, Sora-2: 65.6\%, than in Natural Sciences, Sora-2: \textit{57.7\%}. This divergence suggests that the rigid, immutable laws of physics governing the real world may present a more formidable challenge for models to internalize than the patterns of social dynamics.

\vspace{1mm}

\textbf{4. Advanced Mimicry vs. True Understanding.}
By juxtaposing Table~\ref{tab:explicit_results} and Table~\ref{tab:implicit_results}, we uncover a critical ``Perception-Prediction Gap'' that distinguishes advanced mimicry from true understanding. A striking dissonance is observed in open-weights models like \texttt{Wan-2.2}, which achieves a respectable competency in Implicit Generative Action (\textit{41.5\%}) despite a near-zero score in Explicit Understanding (\textit{0.7\%}). This implies that such models operate via statistical pixel mimicry, rendering the texture of causal events without comprehending the underlying logic. In contrast, \texttt{Sora-2} demonstrates a synchronized profile (Explicit: \textit{51.2\%}, Implicit: \textit{61.6\%}). This correlation suggests that as models scale towards authentic ``World Simulators,'' their generative actions become increasingly grounded in an explicit grasp of causal laws, marking the transition from surface-level synthesis to physically consistent simulation.

\begin{table*}[t]
\caption{Score (\%) on \textit{Implicit Generative Action} Tasks. The \titlered{Red} represents better AVG performance. \vspace{-3mm}}
\setlength{\tabcolsep}{2pt} 
\centering 
\begin{adjustbox}{width=0.99\textwidth}
\begin{tabular}{lccccc|cccccc|c}
\toprule 
\multirow{2.5}{*}{\textbf{Model}} & 
\multicolumn{5}{c}{\textbf{Natural Sciences}} & 
\multicolumn{6}{c}{\textbf{Sociology \& Humanities}} & 
\multirow{2}{*}{\underline{\textcolor{titlered}{\textbf{AVG}}}} \\
\cmidrule(lr){2-6} \cmidrule(lr){7-12}
 & \textbf{Physical.} & \textbf{Natural.} & \textbf{Materials.} & \textbf{Fields.} & \textbf{AVG} & \textbf{Transport.} & \textbf{Life.} & \textbf{Occupat.} & \textbf{Public.} & \textbf{Sports.} & \textbf{AVG} & \\
\midrule
\rowcolor[rgb]{ .961, .985, .970 }
\text{Wan-2.2-14B~\citep{wan2025wan}} & 52.1 & 27.0 & 48.5 & 49.5 & 38.6 & 51.1 & 29.7 & 33.3 & 30.0 & 46.0 & 44.3 & 41.5 \\
\rowcolor[rgb]{ .961, .985, .970 }
\text{Hunyuan-1.5~\citep{wu2025hunyuanvideo}} & 44.4 & 33.8 & 39.8 & 47.7 & 41.2 & 42.9 & 46.0 & 48.4 & 34.0 & 34.5 & 44.0 & 42.5 \\
\rowcolor[rgb]{ .961, .985, .970 }
\text{Veo-3.1~\citep{veo3}} & 60.6 & 58.1 & 51.6 & 54.0 & 57.0 & 49.1 & 69.8 & 67.4 & 61.3 & 55.3 & 60.8 & 58.9 \\
\rowcolor[rgb]{ .961, .985, .970 }
\text{Sora-2~\citep{sora2}} & 56.9 & 59.0 & 59.3 & 55.7 & \textbf{57.7} & 60.6 & 73.0 & 65.1 & 69.8 & 52.0 & \textbf{65.6} & \underline{\textcolor{titlered}{\textbf{61.6}}} \\
\bottomrule 
\vspace{-6mm}
\end{tabular}
\end{adjustbox}
\label{tab:implicit_results}
\end{table*}

\subsection{Causal–Generative Dual-Judge Analysis}

To gain deeper insights into the underlying mechanisms of the ``Perception-Prediction Gap'' and address potential formatting confounders, we conduct a series of in-depth analytical experiments.

\vspace{2mm}

\begin{figure}[t]
	\centering
  \includegraphics[width=\linewidth]{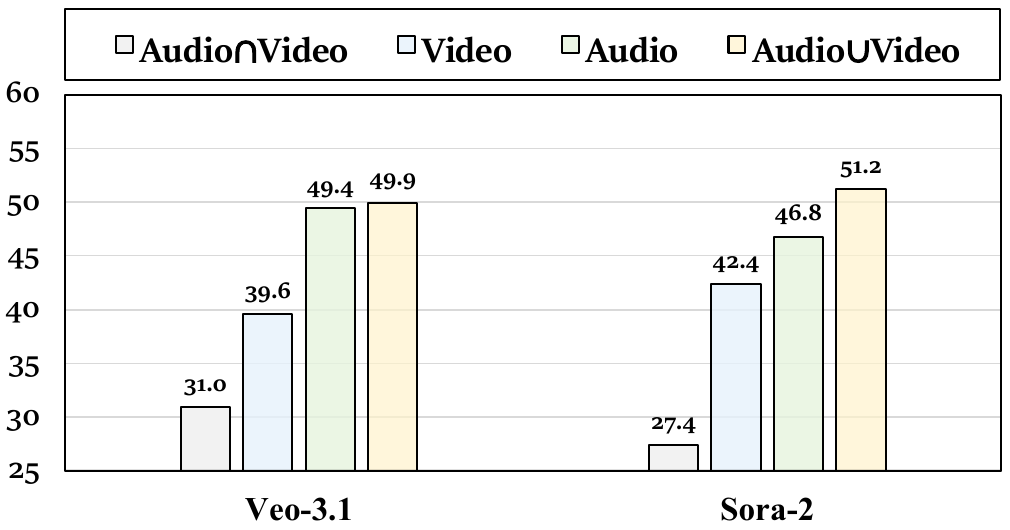}
  \caption{Comparison of \textit{Audio $\cap$ Video}, \textit{Video-Only} and \textit{Audio-Only}. The empirical results show that Audio accuracy consistently surpasses Video accuracy by a significant margin. This indicates models can semantically describe outcomes but fail to physically generate them. \vspace{-3mm}}
  \label{fig_only}
  \vspace{-2mm}
\end{figure}

\textbf{1. The Disconnect Between Linguistic Knowledge and Generative Action.} 
To deeply diagnose the ``Perception-Prediction Gap'', we evaluate the consistency between the model's verbal output and visual output under \textit{identical input conditions}. Specifically, we assess the generation accuracy across two distinct channels: the \textit{Audio-Only} (whether the model verbally articulates the correct reasoning/answer) and the \textit{Video-Only} (whether the model physically renders the correct reasoning). 
The results shown in Figure~\ref{fig_only}, the comparative results reveal a striking Audio-Visual Misalignment independent of instruction-following recognition accuracy. We observe that the accuracy of Audio generation significantly surpasses that of Video generation. This indicates that the model has successfully internalized the \textit{Semantic Knowledge} necessary to describe the causal outcome, proving it theoretically ``understands'' the scenario. However, this semantic comprehension fails to transfer into \textit{Physical Grounding}. The model acts as a ``theoretical expert but practical novice'', capable of verbally predicting the future but incapable of simulating the pixel-level dynamics. This result demonstrates that current models rely on linguistic priors to solve problems, rather than operating on a robust internal world model.

\vspace{2mm}

\begin{figure}[t]
	\centering
  \includegraphics[width=\linewidth]{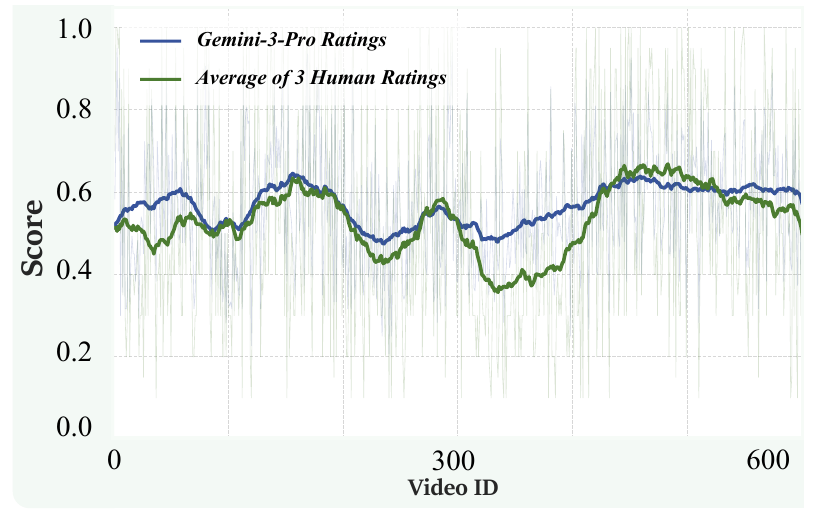}
  \caption{Alignment analysis between \texttt{Gemini-3-Pro Ratings} (Blue) and \textit{Human Expert Ratings} (Green) on the Implicit Generative Action task. The plot illustrates the score distribution across video samples. With a high Pearson Correlation of \textit{0.82} and a low Mean Absolute Difference of \textit{0.14}.}
  \label{fig_human_align}
\end{figure}

\textbf{2. Validation of Automated Scoring via Human Alignment.}
To validate the reliability of our \texttt{Gemini-3-Pro} evaluator, we conduct a Human Evaluation Alignment study. Three human experts blindly rate the test set, and we compute their average scores to benchmark physical and consistency. The result, as shown in Figure~\ref{fig_human_align}, comparing their consensus scores against the model's predictions reveals a high degree of concordance. We achieve a \textit{Pearson Correlation} of \textit{0.8205} and a \textit{Mean Absolute Difference} of \textit{0.1468}. These metrics confirm that \texttt{Gemini-3-Pro} mirrors human judgment, serving as a robust agent for evaluation.

\vspace{2mm}

\textbf{3. The Trade-off Between Temporal Density and Spatial Resolution.} 
To validate the optimal sampling strategy for the \textit{Flatten Temporal Video}, we evaluate the reasoning capacity under varying temporal granularities. The results are as follows:

\vspace{2mm}

\begin{figure*}[t]
	\centering
  \includegraphics[width=\textwidth]{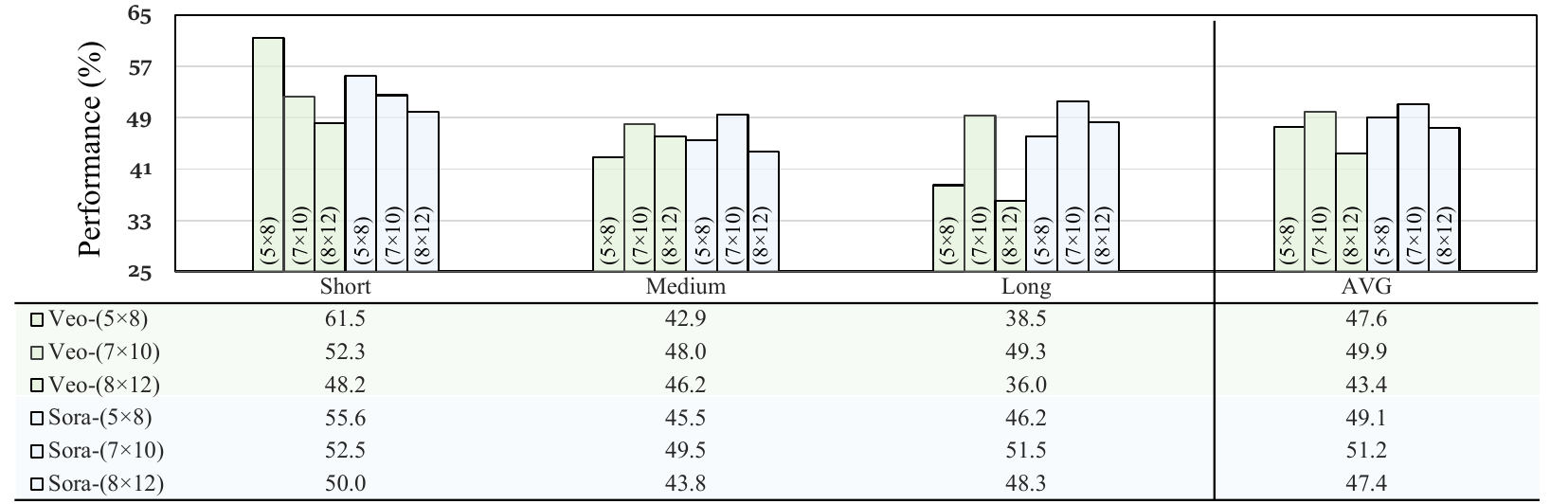}
  \vspace{-6mm}
  \caption{\textbf{Performance comparison across different spatial grid layouts and video durations.} The bar chart and table illustrate the Explicit Causal Understanding accuracy for \texttt{Veo-3.1} and \texttt{Sora-2} under three specific flattening strategies: $5\times8$ (sparse), $7\times10$ (balanced), and $8\times12$ (dense). The $7\times10$ setting yields the optimal average performance across all categories.}
  \label{fig_detail_data}
\end{figure*}

\begin{figure}[t]
	\centering
  \includegraphics[width=\linewidth]{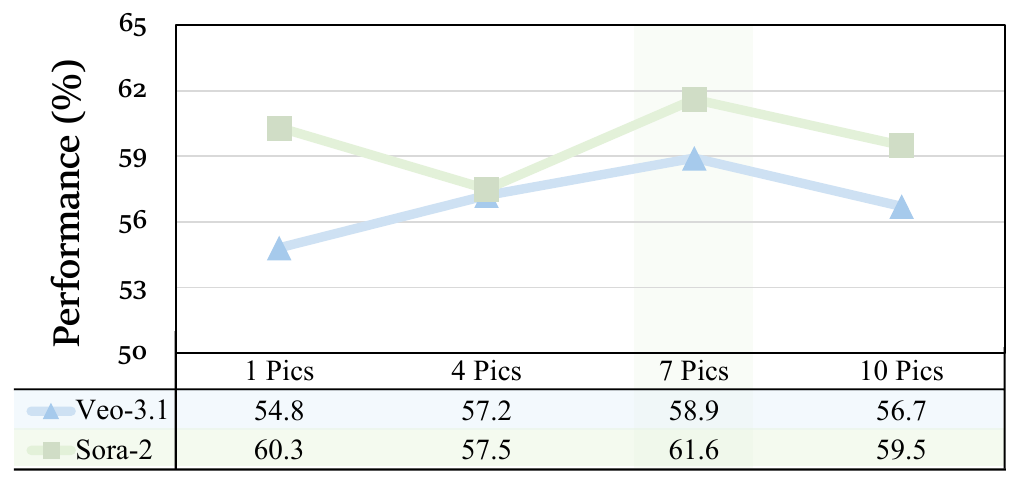}
  \vspace{-5mm}
  \caption{Performance comparison across varying inputs for the action task. The results demonstrate that the 7-Pics yields the highest fidelity, outperforming both sparse and dense contexts by achieving an optimal balance.}
  \label{fig_pics}
  \vspace{-3mm}
\end{figure}

\begin{enumerate}
\item [\ding{182}] \textbf{Explicit Causal Understanding (5$\times$8 vs. 7$\times$10 vs. 8$\times$12):} 
    As shown in Figure~\ref{fig_detail_data}, the results reveal a distinct inverted-U performance curve across most durations, peaking at our default $7\times10$ configuration. Notably, increasing density to $8\times12$ causes a performance drop (e.g., Veo-3.1 falls from $49.9\%$ to $43.4\%$, while Sora-2 drops to $47.4\%$). This suggests that while more frames theoretically offer richer temporal context, they induce severe \textit{Visual Compression} that hinders semantic extraction from the dense grid due to reduced cell resolution. Furthermore, while the $5\times8$ layout benefits short clips, it lacks the necessary temporal resolution required for complex causal reasoning in longer sequences.
\vspace{2mm}
\item [\ding{183}] \textbf{Implicit Generative Action (1 vs. 4 vs. 7 vs. 10 Pics):} 
    As shown in Figure~\ref{fig_pics}, comparing the results confirms that 7-pics represent the optimal prompt, outperforming both sparse (1, 4 pics) and dense (10 pics) inputs. The lower scores at fewer frames suggest \textit{Motion Ambiguity}, where the model lacks sufficient temporal vectors to extrapolate future states. Conversely, the slight decline at 10 frames points to \textit{Contextual Redundancy} where highly compressed keyframes introduce spatial noise that overwhelms the visual encoder's semantic feature extraction.
\end{enumerate}
In summary, these findings confirm that our chosen configurations ($7\times10$ grid and 7 frames) strike the optimal balance between preserving spatial fidelity and encoding sufficient temporal dynamics required for robust simulation.

\begin{section}{Related Work}\label{sec:related}
\textbf{Towards ``Thinking with Video''.} 
Recent advancements in Video Generation Model have evolved beyond high-fidelity texture synthesis toward complex spatiotemporal dynamics modeling~\cite{zhou2025scaling,lu2025rewardforcingefficientstreaming,li2025stable}. This progression supports the conceptualization of video generators as nascent general world simulators and promotes the ``thinking with video'' paradigm, where generation functions as an intrinsic component of reasoning~\cite{tong2025thinking,yang2025longvt,luo2025vreasonbenchunifiedreasoningbenchmark,zhang2025vitcot}. However, distinguishing genuine causal understanding from mere statistical memorization remains a pivotal challenge, as traditional metrics disproportionately favor perceptual fidelity over logical consistency.

\textbf{The Gap in Existing Reasoning Benchmarks.} 
Emerging benchmarks evaluate various reasoning facets, ranging from rule-based dynamics~\cite{he2025rulerbenchprobingrulebasedreasoning,liu2025worldsimulatorsreasongenvire,zhang-etal-2025-cchall} to open-ended visual reasoning~\cite{qin2026large,chen2025tivibench,luo2025vreasonbenchunifiedreasoningbenchmark} and specific tasks~\cite{yang2025reasoning}. While recent efforts attempt to mitigate appearance biases~\cite{guo2025video,deng2025video,zhang2026latent,cai2025mmgr,li2026frameprocessawareevaluationgenerative}. However, relying on rudimentary tasks, they fail to verify physical and social dynamics. By evaluating discrimination and generation in isolation, they overlook the gap between surface-level outputs and genuine causal understanding.

To address the critical gap in current causal evaluation and in contrast to existing methods, we introduce the \textit{Causal–Generative Dual-Judge (\texttt{CGDJ})} framework. Unlike previous works, \texttt{CGDJ} jointly evaluates internal consistency through two complementary modalities, \textit{Explicit Causal Understanding} and \textit{Implicit Generative Action}. By constructing the Causal–Generative Evaluation Benchmark, we provide a rigorous testbed to distinguish whether video generative models have genuinely internalized the physical laws and causal structures governing the scenes they render.

\end{section}

\begin{section}{Conclusions}

We redefine reasoning with video generative models as \textbf{Thinking in Video} and introduce the \textbf{Causal-Generative Dual-Judge (\texttt{CGDJ})} to evaluate whether current generators can support this paradigm. By jointly examining \textit{Explicit Causal Perception} and \textit{Implicit Generative Prediction}, we reveal a clear \textbf{Perception-Prediction Gap}: models may generate plausible visual dynamics without reliable causal perception, while stronger systems show early but limited alignment between reasoning and generation. We further identify audio-visual misalignment, where verbalized causal knowledge does not necessarily translate into consistent video rollouts. Ultimately, \texttt{CGDJ} advocates shifting video generation evaluation from perceptual fidelity toward causal consistency, offering diagnostic insights for grounded, reasoning-capable world models.

\end{section}

\balance

\bibliographystyle{unsrtnat}
\bibliography{sample-base}

\end{document}